\documentclass[review]{elsarticle}

\usepackage{amsmath}

\usepackage[dvipsnames]{xcolor}

\usepackage{float}

\usepackage{graphicx} 
\graphicspath{{images/}}
\DeclareGraphicsExtensions{.eps,.pdf,.jpg,.png}

\usepackage[caption=false]{subfig}

\usepackage{multicol}
\usepackage{multirow}

\usepackage{url}

\usepackage{verbatim}

\usepackage{tabularx} 
\usepackage{booktabs}
\usepackage{array}
\usepackage{siunitx} 

\usepackage{color}

\newcommand{\af}[1]{{\color{black}#1}}
\newcommand{\nrh}[1]{{\color{black}#1}}

\newcommand{\old}[1]{{#1}}
\newcommand{\x}[0]{\phantom{0}}

\def\bbbr{{\rm I\!R}}
\newcommand{\ged}{\operatorname{GED}}
\newcommand{\hed}{\operatorname{HED}}
\newcommand{\bp}{\operatorname{BP}}
\DeclareMathOperator*{\avg}{\operatorname{avg}}

\usepackage{adjustbox}
\usepackage{pdflscape}

\newcommand{\new}[1]{{\color{black}#1}}
\journal{Pattern Recognition}

\begin{document}

\begin{frontmatter}


\title{\af{Graph-Based Offline Signature Verification}}


\author[diuf]{Paul Maergner}
\author[smith]{Nicholas R. Howe}
\author[northwestern]{Kaspar Riesen}
\author[diuf]{Rolf Ingold}
\author[diuf,heia]{\\Andreas Fischer}

\address[diuf]{University of Fribourg, Department of Informatics, DIVA Group, Fribourg, Switzerland}
\address[smith]{Smith College, Department of Computer Science, Northampton, Massachusetts, USA}
\address[northwestern]{University of Applied Sciences and Arts Northwestern Switzerland, Institute for Information Systems, Olten, Switzerland}
\address[heia]{University of Applied Sciences and Arts Western Switzerland, Institute of Complex Systems, Fribourg, Switzerland}


%

\begin{abstract}

Graphs provide a powerful representation formalism that offers great promise to benefit tasks like handwritten signature verification.
While most state-of-the-art approaches to signature verification rely on fixed-size representations, graphs are flexible in size and allow modeling local features as well as the global structure of the handwriting.
In this article, we present two recent graph-based approaches to offline signature verification: keypoint graphs with approximated graph edit distance and inkball models. 
We provide a comprehensive description of the methods, propose improvements both in terms of computational time and accuracy, and report experimental results for four benchmark datasets. 
The proposed methods achieve top results for several benchmarks, highlighting the potential of graph-based signature verification.

\end{abstract}


\end{frontmatter}

%

\section{Introduction}

Handwritten signatures are commonly used for personal authentication in many areas.
Along with the broad use of signatures comes an interest in the verification of their authenticity.
Signature verification is a challenging task even for humans since the decision has to be made using only a few genuine samples.
The development of automatic signature verification systems started decades ago and remains an active field of research~\cite{Hafemann2017review,diaz2019}

Two different cases of signature verification are commonly distinguished; viz.~\emph{online} (also termed dynamic) and \emph{offline} (also termed static) signature verification.
The former leverages dynamic characteristics like speed, timing, and pressure.
This also implies that the signatures have to be collected with a special device like a pen tablet or a touchscreen.
\emph{Offline} signature verification, on the other hand, uses only the image of the signature, e.g.~a scan of a signature written on paper.
Overall, the offline case can be applied to more use cases, but it is also considered the more challenging task due to the lack of additional information.
The work presented in this paper \nrh{considers} the offline case.

The majority of state-of-the-art offline signature verification systems rely on statistical pattern recognition, i.e.~handwritten signatures are represented using fixed-size feature vectors (or sequences of feature vectors).
The feature vectors either rely on manually engineered features or more recently on features automatically learned from signature images using neural networks.
The engineered features commonly represent either local information such as local binary patterns (LBP) and histogram of oriented gradients (HOG)~\cite{Yilmaz2011OfflineFeatures}, or Gaussian grid features taken from signature contours~\cite{Nguyen2011AnVerification}, or they represent global information, for example, geometrical features like Fourier descriptors, number of holes, number of branches in the skeleton, moments, projections, distributions, position of barycenter, tortuosities, directions, curvatures and chain codes, and many others \cite{Vargas2011,Nguyen2007Off-lineMachines,Armand2006Off-lineFeature,Hassaine2011TheContest,Gilperez2008Off-lineFeatures,Impedovo2008,Plamondon1989}.
With the advances in deep learning, we can see a shift from handcrafted features towards learned features\af{, for example by} using deep convolutional neural networks (CNN)~\cite{Hafemann2017}.
In both cases, engineered as well as learned features, the fixed-size feature vectors are then used in conjunction with statistical classifiers or matching schemes that operate on statistical streams, e.g.~hidden Markov models (HMM), support vector machines (SVM)~\cite{Justino2005AVerification,Ferrer2005}, dynamic time warping (DTW)~\cite{PiyushShanker2007}\new{, or neural networks~\cite{Soleimani2016patreclet}.}

While most state-of-the-art systems for signature verification rely on feature vectors and statistical classifiers, there exists a more powerful representation formalism: graphs.
\nrh{Not only can they} model a variable amount of information \af{by means of} nodes but \nrh{they} also directly model binary relations between them \af{by means of} edges.
When used to represent signatures, nodes usually represent elementary strokes of the handwriting or keypoints in the signature images.
Edges model the relations between these parts in the global structure.
\nrh{Considering the vast number of ways that graphs can be used to represent signatures, it is somewhat surprising that they have only rarely been applied to signature verification in the past.
One of the main reasons may be the typically high computational complexity of graph operations such as graph matching.}
Previous works include the early proposal by Sabourin et al.~to represent signatures based on stroke primitives~\cite{sabourin94structural}, the proposal by Bansal et al.~to use a modular graph matching approach~\cite{Bansal2009}, and the proposal by Fotak et al.~to use basic concepts of graph theory~\cite{Fotak2011}.

A more recent approach for graph-based signature verification has been proposed by Maergner et al.~\cite{maergner2017icdar}.
The introduced framework is based on the graph edit distance between labeled graphs representing individual signatures.
In the experiments, \af{keypoint graphs have been used}, which have also been \af{considered} for handwriting recognition~\cite{Fischer2010} and keyword spotting~\cite{Stauffer2016} in the past.
\af{The well-known bipartite approximation~\cite{Riesen2009} of graph edit distance has been employed} to reduce the high computational complexity of matching two graphs.

Inkball models are another recent structural approach for handwriting analysis proposed by Howe in~\cite{howe:icdar2013}.
This approach has been introduced as a technique for segmentation-free word spotting that requires \af{few} training data.
In addition to keyword spotting, inkball models have been used for handwriting recognition \af{as a complex feature in conjunction} with HMM~\cite{howe:icfhr2016}.
Inkball models are visually similar to keypoint graphs since they are using very similar points on the handwriting as nodes.
But the inkballs are connected to a rooted tree that is directly matched with a skeleton image using an efficient algorithm.

\af{In a preliminary work~\cite{maergner2018icfhr}, we introduced an inkball-based signature verification system.
The method was evaluated individually as well as combined with the graph edit distance based approach, demonstrating a promising signature verification performance on two benchmark datasets.}

\af{In the present article, we continue this line of research and extend our previous work as follows.
First, we provide a more thorough description of the two graph-based signature verification systems, i.e.~one based on graph edit distance and the other based on inkball models.
Secondly, we propose improvements for both methods.
The graph edit distance approach is rendered more efficient by using the quadratic-time Hausdorff edit distance~\cite{Fischer2015} instead of the cubic-time bipartite approximation~\cite{Riesen2009}.
The inkball approach is rendered more accurate by means of an augmented inkball model that is introduced in the present article.
It considers angular information in addition to the inkball position.
Thirdly, we conduct an extensive experimental evaluation of both methods on four publicly available benchmark datasets, in order to compare them with the current state of the art.
The systems are evaluated individually as well as combined, to profit from their complementary properties.
We demonstrate that graph-based signature verification is able to reach and, in some cases, surpass the current state of the art in signature verification, motivating further research on structural approaches to signature verification.}


This paper is structured in the following way.
In Section~\ref{sec:graph}, keypoint graphs, the graph matching approach, and the dissimilarity based on graph edit distance are formally \af{introduced}.
In Section~\ref{sec:inkball}, inkball models, the augmented matching approach, and the inkball dissimilarity are described.
In Section~\ref{sec:sigver}, we discuss how we use the two dissimilarity models individually as well as in combination for offline signature verification.
In Section~\ref{sec:exp}, we present and discuss our experimental results.
Finally, we draw conclusions in Section~\ref{sec:concl} and provide pointers to future lines of research.

%

\section{Graph Edit Distance \label{sec:graph}}

Our first structural \af{approach to signature verification} is based on the graph edit distance (GED). 
Initially, this approach has been introduced in~\cite{maergner2017igs} and was further refined in~\cite{maergner2017icdar}.
The main idea is that signature images are represented using graphs and these graphs are then compared using GED.
GED is one of the most flexible graph matching approaches available since it can be used to match any kind of labeled graph given an appropriate cost function.
We are using keypoint graphs as our graph representation.
As an approximation for GED, we consider the Hausdorff edit distance~\cite{Fischer2015} as opposed to the bipartite approximation that has been used in previous publications~\cite{maergner2017icdar,maergner2018icfhr,maergner2018ssspr}.
In the following sections, the individual building blocks of this approach are described in more detail.

\subsection{Image Processing and Graph Representation}
\begin{figure*}[t]
\centering
\includegraphics[width=\textwidth,trim={12mm 10mm 0mm 0mm},clip]{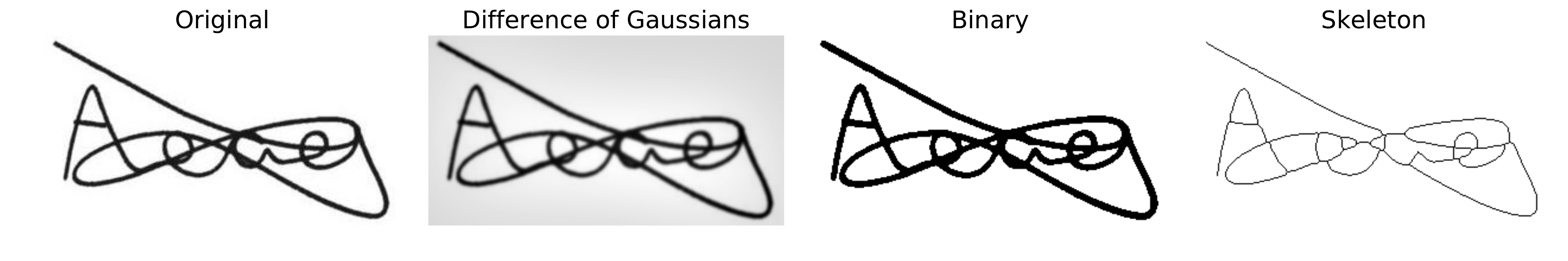}
\caption{Image processing for keypoint graph shown on first signature of user 3902 from the GPDSsynthetic dataset~\cite{Ferrer2015}. \label{fig:imageprocessing}}
\end{figure*}

Keypoint graphs are built from a skeleton image of handwriting. 
Thus, the signature images have to be binarized and skeletonized first. 
This is achieved using the following steps:
\begin{itemize}
    \item A local edge enhancement is performed using a difference of Gaussian (DoG) filter on grayscale signature images.
    \item The enhanced image is binarized using a global threshold.
    \item The binary image is finally thinned to single pixel width using the algorithm proposed in~\cite{Zhang1984APatterns}.
\end{itemize}
The three preprocessing steps are visualized in Fig.~\ref{fig:imageprocessing}.

A labeled graph $g$ is defined as a four-tuple $g = (V,E,\mu,\nu)$, where 
$V$ is the finite set of nodes,
$E\subseteq V \times V$ is the set of edges,
$\mu:V\rightarrow L_V$ is the node labeling function, and
$\nu: E \rightarrow L_E$ is the edge labeling function.

In keypoint graphs, the nodes represent keypoints on the handwriting and the node labels are the coordinates of these points, i.e. $L_V = \bbbr \times \bbbr$.
The edges are unlabeled and undirected, i.e. $L_E = \emptyset$ and $(u,v) \in E \iff (v,u) \in E$, and connect two nodes if their corresponding points are directly connected on the handwriting.

\begin{figure}[t]
\centering
\includegraphics[width=0.8\textwidth,trim={10mm 10mm 0mm 8mm},clip]{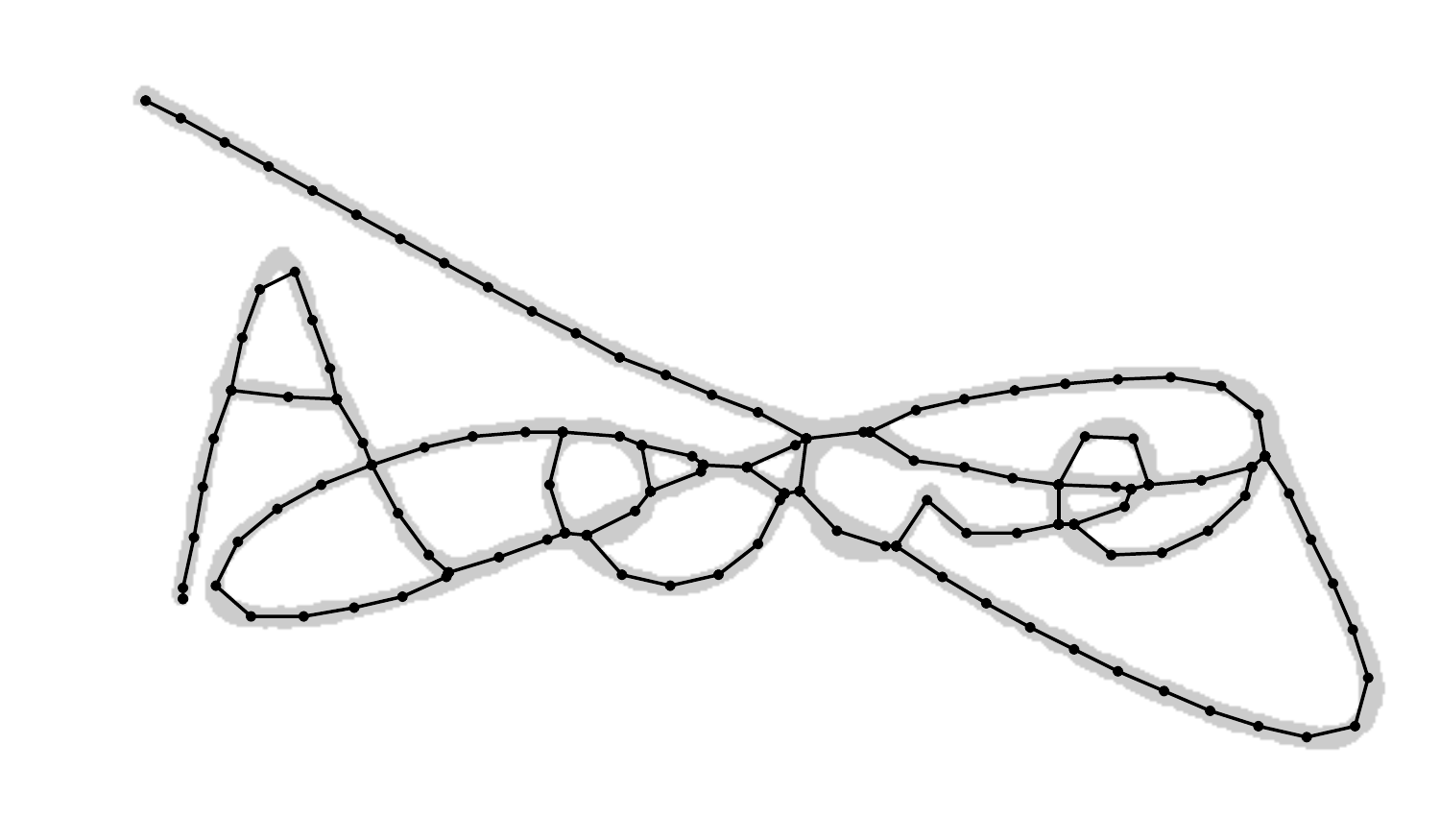}
\caption{Example keypoint graph generated from the first signature of user 3902 from the GPDSsynthetic dataset~\cite{Ferrer2015} \label{fig:graph}}
\end{figure}

Specifically, the nodes and edges are extracted from the skeleton image of the signature.
The keypoints are selected iteratively.
First, junction-points and end-points are added to the set of keypoints. 
Secondly, the left outer most pixel of circular structures is added to the keypoints if they contain no keypoints, yet.
Then, additional points are added by sampling the skeleton.
This is done by tracing along the skeleton while starting at already selected keypoints. 
Once the traveled distance without meeting a keypoint is larger or equal to $D_\text{GED}$, a new keypoint is added.
While tracing along the skeleton, we also place edges to connect neighboring keypoints.
The node labels are finally normalized to make the graph representation translation invariant by subtracting the average node label from each node label in the graph. Thus, the nodes in the graph are centered around the origin $(0,0)$ in the two dimensional plane.
An example of a keypoint graph is shown in Fig.~\ref{fig:graph}.
In this paper, we call a keypoint graph $g_R$ if it is based on a signature image $R$.

\subsection{Edit Distance}

Graph edit distance (GED) offers one of the most flexible ways to measure the dissimilarity between two labeled graphs~\cite{Bunke1983InexactRecognition,Riesen2015StructuralDistance}.
It calculates the cost of the minimal cost transformation of graph $g_R = (V_R,E_R,\mu_R,\nu_R)$ into graph $g_T = (V_T,E_T,\mu_T,\nu_T)$.
Hereby, a valid transformation is a sequence of edit operations that completely transform graph $g_R$ into graph $g_T$.
Commonly, the edit operations are defined as substitution, deletions, and insertions of both nodes and edges.
An appropriate cost function defines the cost of each of these edit operations.
This allows GED not only to handle any kind of labeled graphs but also to be adapted to a specific task using tailored cost functions.
Unfortunately, exact computation of GED have a major disadvantage: its computational complexity.
Exact computation of GED belongs to the group of NP-hard problems and its complexity is exponential in the number of nodes \af{in} the two graphs.
In practice, this means that exact GED can be calculated within reasonable time only for rather small graphs.

To be able to use the concept of GED, we have to use an approximation for GED.
A very popular approximation of GED is the bipartite approximation\footnote{Sometimes also referred to as assignment edit distance (AED).} (BP) proposed by Riesen and Bunke~\cite{Riesen2009}. 
BP reduces the problem of GED to an instance of a linear sum assignment problem (LSAP) and thus has a cubic time complexity, specifically, it is in $O((n+m)^3)$, where $n=|V_R|$ and $m=|V_T|$ are the number of nodes in the two graphs. 
We used BP in our previous publications~\cite{maergner2017icdar,maergner2018icfhr,maergner2018ssspr}.
In the present article, we are going to use a more efficient approximation based on Hausdorff edit distance (HED)~\cite{Fischer2015}, which operates in quadratic time complexity, specifically, $O(n \cdot m)$. 
The method is described in more detail in the next section.

\subsection{Hausdorff Edit Distance}

The Hausdorff Edit Distance (HED) proposed by Fischer et al.~\cite{Fischer2015} computes a lower bound of GED, i.e. it is always smaller than or equal to GED. 
The main idea is to find locally optimal assignments between substructures of one graph and substructures of the other graph. 
For HED, such substructures consist of nodes together with their adjacent edges.
Formally, the HED of two graphs $g_R$ and $g_T$ given a cost function $c$ is defined as
\begin{equation}
	\begin{split}
	\hed(g_R, g_T, c) = \sum_{u \in V_R}\min_{v \in V_T \cup \{\varepsilon\}}c^*(u \rightarrow v) + \sum_{v \in V_T}\min_{u \in V_R \cup \{\varepsilon\}}c^*(u \rightarrow v)
	\end{split}
\end{equation}
\af{Where $c^*(u \rightarrow v)$ takes into account the cost of substituting node $u$ with node $v$ as well as matching the adjacent edges of $u$ with the adjacent edges of $v$.}
Similar to the Hausdorff distance, this assignment is performed in both directions: all nodes of $g_R$ are assigned to nodes of $g_T$ or $\varepsilon$ \af{(indicating deletion or insertion)}, and vice versa.

\subsection{Cost Function \label{sec:graph:cost}}
To compute an edit distance, we have to define an appropriate cost function $c$ for the chosen graph representation.
The cost function $c$ assigns a cost to substitutions, deletions, and insertions of both nodes and edges.

In our scenario, we set the node substitution cost to the Euclidean distance between the node labels of $u$ and $v$:
\begin{equation}
c(u \rightarrow v) = \sqrt{(x_u - x_v)^2+(y_u - y_v)^2} \text{,}
\end{equation}
where $\mu_R(u) = (x_u, y_u)$ and $\mu_T(v) = (x_v, y_v)$ are the node labels of nodes $u \in V_R$ and $v \in V_T$, respectively.

We use a constant cost $c_\text{node}$ for both deletions and insertions of nodes:
\begin{equation}
c(u \rightarrow \varepsilon) = c(\varepsilon \rightarrow v) = c_\text{node}
\end{equation}

The edge substitution cost is set to zero:
\begin{equation}
c(e_R \rightarrow e_T) = 0 \text{,}
\end{equation}
where $e_R \in E_R$ and $e_T \in E_T$.

The cost of edge deletion and insertion is set to a constant value $c_\text{edge}$.
\begin{equation}
c(e_R \rightarrow \varepsilon) = c(\varepsilon \rightarrow e_T) = c_\text{edge}
\end{equation}

\subsection{Normalization of Graph Edit Distance}
\af{In the following,} we use $\ged(\cdot)$ as a placeholder for a GED approximation, i.e. $\bp(\cdot)$ or $\hed(\cdot)$.
Previous publications have shown that it is crucial to apply a good normalization when using GED for signature verification.
We normalize our GED with what we call "maximal GED", i.e. the cost of deleting the first graph and then inserting the second graph.

Formally, given two graphs $g_R = (V_R,E_R,\mu_R,\nu_R)$ and $g_T = (V_T,E_T,\mu_T,\nu_T)$ and a cost function $c$, we define $\ged_\text{max}$ as
\begin{equation}
    \begin{split}
    \ged_\text{max}(g_R, g_T) = \sum_{u \in V_R}{c(u \rightarrow \varepsilon)} + \sum_{e \in E_R}{c(e \rightarrow \varepsilon)} \\
                                + \sum_{v \in V_T}{c(\varepsilon \rightarrow v)} + \sum_{e \in E_T}{c(\varepsilon \rightarrow e)}
    \end{split}
\end{equation}
When using the cost function defined in section~\ref{sec:graph:cost}, this equation can be simplified to
\begin{equation}
    \begin{split}
    \ged_\text{max}(g_R, g_T) = (|V_R|+|V_T|) \cdot c_\text{node} + (|E_R|+|E_T|) \cdot c_\text{edge}
    \end{split}
\end{equation}

We now define the normalized GED-based dissimilarity of two signature images $R$ and $T$ as
\begin{equation}
d_{\ged}(R,T) = \frac{\ged(g_R,g_T)}{\ged_{\text{max}}(g_R,g_T)} \text{,}
\end{equation}
where $g_R$ and $g_T$ are the keypoint graphs for the signature images $R$ and $T$ respectively, and $\ged(g_R,g_T)$ is an approximation of the GED using either BP or HED.

\section{Inkball Models}
\label{sec:inkball}
Inkball models are a type of part-structured model \cite{felz:ijcv05} introduced by Howe as a technique for word spotting \cite{howe:icdar2013} but have since been used for matching handwritten symbols of various kinds, recently including signatures \cite{howe:icdar2015,maergner2018icfhr}.  
Inkball models may be visualized as a combination of rigid displacements that give a default shape, with spring potentials at each joint that add the flexibility to match spatially varying versions of the target.  

Like the keypoint graph models discussed above, an inkball model can be built from a single signature image.  
The process is similar, although the details differ mostly thanks to their independent development.  
The signature is first thinned to single pixel width, then inkball nodes are placed at regular intervals along the skeleton.  
The first nodes are placed at all endpoints and junctions.  
Additional nodes are then inserted in a greedy manner on skeleton points at distance $D_{\text inkball}$ from any existing node until such insertion is no longer possible.  
Finally, to fill in any large gaps that might remain, nodes are inserted at locations as far as possible from existing nodes, stopping when the inserted nodes would be less than $\frac{\sqrt{2}}{2}D_{\text inkball}$ from existing nodes.  
(Note that for better presentation the use of $D_{\text inkball}$ \nrh{in this paper} differs from earlier presentations of the inkball method \cite{maergner2018icfhr} by a factor of $\sqrt{2}$; the method itself is unchanged.)

Prior work on inkball matching has not relied on any information about the nodes other than their relative positions.  
This simplicity stands in contrast to much of the body of work on part-structured models, which often incorporate descriptive properties for each part that can be used to improve matching accuracy \cite{felz:pami10}.  
This paper works with the traditional inkball models used by prior work, but also introduces a novel {\em augmented inkball model} that incorporates distinguishing information about each inkball node.  
Specifically, the augmented model records the local tangent angle of the ink skeleton in the neighborhood of the inkball's location.  
This tangent information is computed as follows.  
First, we separate the skeleton into one-dimensional arcs by splitting it at every junction point.  
For each arc, we then smooth the pixel-to-pixel tangent sequence using a 1D Gaussian filter ($\sigma = 2$) to suppress noise.  
Finally, each node takes the tangent angle of its nearest arc point (nodes that represent junctions take the value from one of their connected arcs by random).

Once all the nodes are identified and described, they are linked into a tree structure that allows for efficient matching.  
Linking also takes a greedy approach.
First, connections are added between pairs of the nearest disconnected nodes (w.r.t. their Euclidean distance). 
This process is repeated until the entire set of nodes is connected.  
Next, a single node is designated as the root of the tree.  
By convention the node closest to the centroid is chosen.  
Before any approximations, the matching energy described below is mathematically equivalent regardless of the root choice.

Using a tree structure means that the model cannot fully represent loops in the original signature.  
Each loop will include a break at some arbitrary location, and the model will sometimes allow the nodes on either side of the break to separate when matching.  
This disadvantage is the price of the efficient tree matching algorithm described below.

The term \nrh{``}inkball model" arises from the fact that when the separation between nodes is small enough, balls of ink placed at each node location will overlap to reproduce an approximation of the original handwritten symbol.  
With each connection displaced by sampling from a 2-dimensional normal distribution, the model can generate novel variations on the original form.  
This enables use of a Bayesian inference framework, where likelihood is estimated by computing the probability of the model under a given set of observations.  
As described in the next section, a dynamic programming implementation makes this computation tractable.

\subsection{Inkball Matching}
\label{sec:inkball:match}

\begin{figure*}[t]
\centering
\includegraphics[width=0.9\textwidth,trim={30mm 55mm 25mm 35mm},clip]{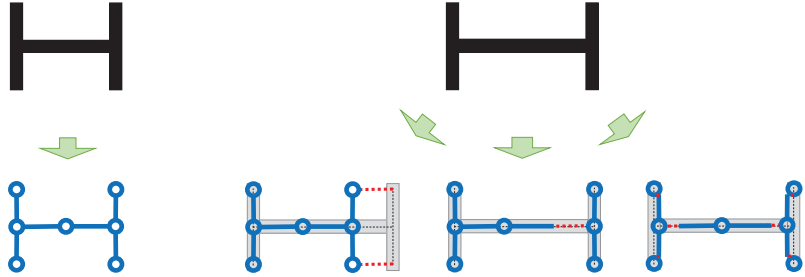}
\caption{Three possible configurations for a hypothetical model (left) matched to an observation (right).  
The lowest energy configuration (far right) places every node near a skeleton point, with small deformations at every link.}
\label{fig:inkballdemo}
\end{figure*}

Inference seeks to maximize the likelihood of the model configuration (e.g., deformation on a sample signature) given an observation (e.g., a test signature).  
Likelihood captures both the placement of inkballs near observed ink and the deformation of the default model displacements required to achieve such a placement.  
The best configuration typically trades off both types of error, and spreads any necessary deformation across the model, as shown in Fig.~\ref{fig:inkballdemo}.  
A match can only be considered plausible if both considerations are well satisfied.  
In the case of augmented inkball models, for a satisfactory likelihood, the nodes must be placed near ink that shows the same or similar tangent angle.

First, it is necessary to establish notational conventions.  
For a given model, assign each node a numeric index, with the root at index 0 by convention.  
A {\em configuration} $C$ is an assignment of 2D positions $\{v_i\}$ to the nodes of the model.  
In any configuration, it is the 2D vector displacements between each node and its parent that are of primary interest.  
These are designated $\vec{s}_i$ and may be computed from the raw configuration positions and a knowledge of the model's connectivity:  
$\vec{s}_i = \vec{v}_{i\uparrow}-\vec{v}_i$.  
(Here $i\uparrow$ indicates the index of the parent of node $i$ \nrh{with respect to the tree structure}.)  
The displacements in the originally observed configuration of the model (as taken from the sample signature) have special significance and are denoted $\vec{o}_i$.  
These are known as the {\em default offsets}, and the configuration that produced them is the {\em rest configuration}.  

Rather than maximizing likelihood directly, it suffices to minimize a related energy function over all possible configurations of the model.  
This energy has two terms, one ($E_{\Omega}$) related to the quality of the node placement relative to the observations, and the other ($E_{\xi}$) related to the deformation implied by that placement.  
In the equations below, $R$ is the sample signature and $T$ is the test signature.

\begin{equation}
E(R,C,T) = E_{\xi}(R,C)+\lambda E_{\Omega}(C,T)
\label{eqn:Ebasic}
\end{equation}
\begin{equation}
E_{\xi}(R,C) = \sum_{i = 2}^n \|\vec{s}_i-\vec{o}_i\|^2
\label{eqn:Espring}
\end{equation}
\begin{equation}
E_{\Omega}(C,T) = \sum_{i = 1}^n \Omega_T(\vec{v}_i)^2
\label{eqn:Edata}
\end{equation}

Equation \ref{eqn:Edata} above relies on an expression $\Omega_t(\vec{v}_i)$ that describes the placement quality of node with index $i$.  
To define it precisely, let $\{\vec{t}_j\}$ represent the 2D vector coordinates of each of the skeleton points in \nrh{$T$}.  

\begin{equation}
\Omega_T(\vec{v}_i) = \min_{j} \delta_{ij}\|\vec{t}_j-\vec{v}_i\|
\label{eqn:Omegadef}
\end{equation}

For the ordinary inkball match, $\delta_{ij} = 1$ in all cases.  
For the augmented match, it depends on a comparison of the local tangent angles, $\alpha_i$ and $\beta_j$.

\begin{equation}
\delta_{ij} = w_{\alpha}\min((\alpha_i-\beta_j) \bmod \pi,(\beta_j-\alpha_i) \bmod \pi)
\label{eqn:deltaij}
\end{equation}

Equation~\ref{eqn:Ebasic} can be minimized efficiently via dynamic programming by organizing the sums of Equations~\ref{eqn:Espring} and \ref{eqn:Edata} according to the nodes involved.  
Let $C_{(i)}$ refer to the configuration of the model subtree rooted at node at index $i$, and define $E^*_{(i)}(\vec{v})$ as a functional giving the minimal energy on that subtree for configurations satisfying $\vec{v}_i = \vec{v}$.

\begin{equation}
E^*_{(i)}(\vec{v}) = \min_{C_{(i)}|\vec{v}_i = \vec{v}} E(R,C_{(i)},T)
\label{eqn:Estari}
\end{equation}

The terms of $E^*_{(i)}(\vec{v})$ can be arranged to compute the desired energy in terms of similar expressions on the children of node at index $i$. 
(The set of their indices is denoted $i\downarrow$ below).

\begin{equation}
E^*_{(i)}(\vec{v}) = \Omega_T(\vec{v})^2 +\sum_{j\in i\downarrow}\left[\min_{\vec{u}} E^{**}_{(j)}\right]
\label{eqn:Eincrement}
\end{equation}

\begin{equation}
E^{**}_{(j)} = \|(\vec{v}-\vec{u})-\vec{o}_i\|^2+E^*_{(j)}(\vec{u})
\label{eqn:Estarstarj}
\end{equation}

Equation~\ref{eqn:Eincrement} is trivial to compute for leaf nodes of the model \nrh{since the summation has no terms}.  
Starting with the leaves, their parent nodes can be computed next, and so on until the root node energy functional $E^*_{(0)}(\vec{v})$ is found.  
Each functional on $\vec{v}$ is represented discretely on a pixel-resolution grid, and the minimization in Equation~\ref{eqn:Eincrement} is computed using the generalized distance transform \cite{felz:ijcv05}.

In many cases, the quadratic error of Equations~\ref{eqn:Espring} and \ref{eqn:Edata} places too much emphasis on badly mismatched nodes.  
A form of truncated quadratic serves better in practice, and is used in the experiments.  

\begin{equation}
{E^*}'_{(i)}(\vec{v}) = \min(E^*_{(i)}(\vec{v}),N_i\tau)
\label{eqn:Etrunc}
\end{equation}

Here $N_i$ represents the number of nodes in the subtree rooted at index $i$ and $\tau$ is a free parameter optimized by experiment.

\begin{figure*}[t]
\centering
\begin{tabular}{cc}
\includegraphics[width=.48\textwidth,trim={0mm 0mm 0mm 0mm},clip]{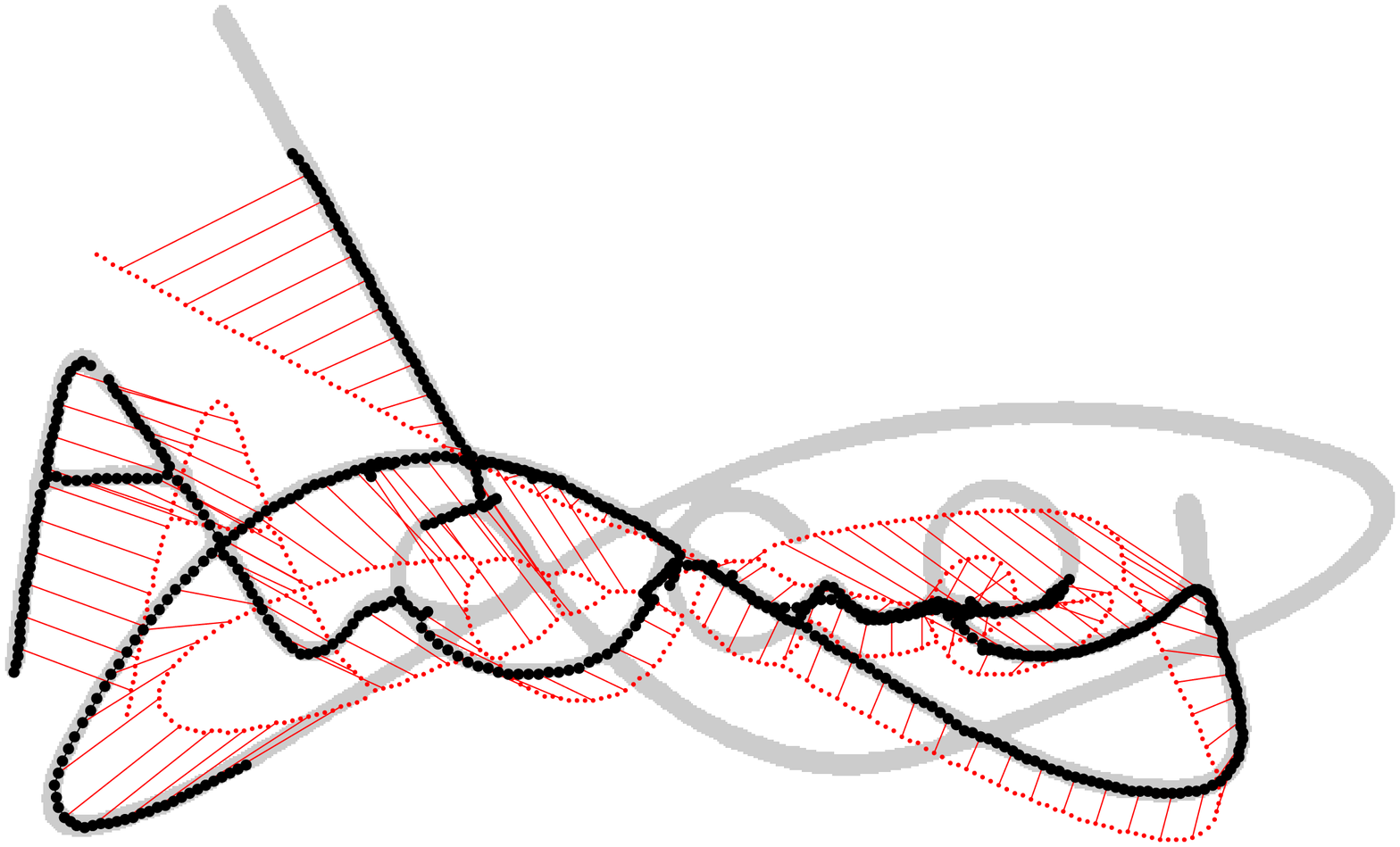}
\includegraphics[width=.48\textwidth,trim={0mm 0mm 0mm 0mm},clip]{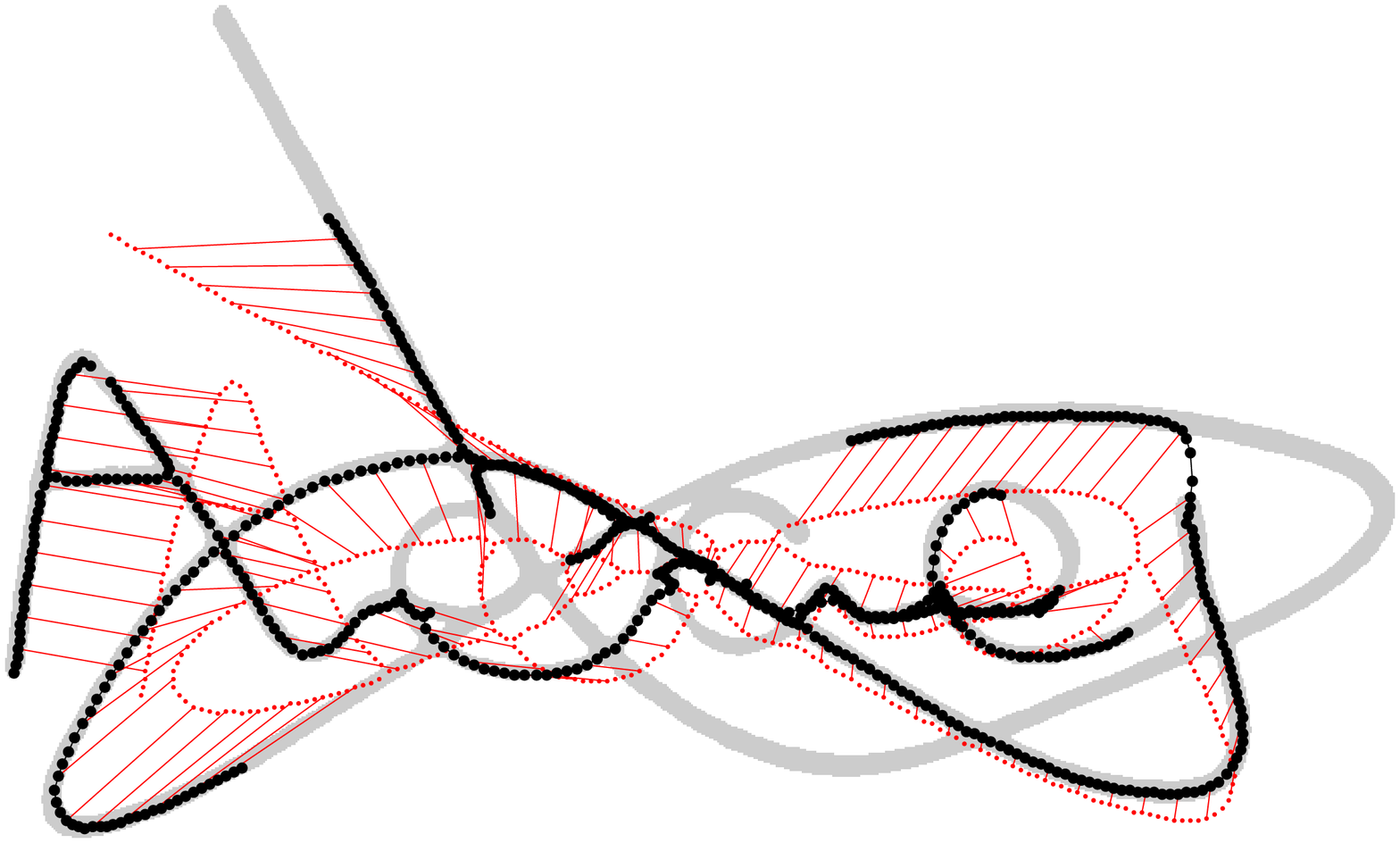}\end{tabular}
\caption{Sample matches of a model to an observation.  
Left is the original inkball match; right is the augmented match.}
\label{fig:inkballmatch}
\end{figure*}

The tools developed above allow the comparison of two signature images $R$ and $T$.  
The former is converted to an inkball model and the latter becomes the observation to be compared against as illustrated in Figure~\ref{fig:inkballmatch}.

To ensure scale invariance it is important to normalize the minimal energy by the number of inkballs in the model ($N_0$), yielding the average energy per inkball.  
This normalization is similar to that applied to the graph edit distance.  
Putting everything together, the inkball dissimilarity of $R$ and $T$ is given by Equation~\ref{eqn:Dinkball} below.

\begin{equation}
d_{\text{inkball}}(R,T) = \frac{1}{N_0} \min_{\vec{v}} {E^*}'_{(0)}(\vec{v})
\label{eqn:Dinkball}
\end{equation}

%

\section{Signature Verification System}\label{sec:sigver}

In our signature verification system, the decision of whether an unseen signature is a genuine signature of the claimed user is made based on a set $\mathcal{R}$ of known genuine signatures from that user, termed {\emph references}.
An unseen signature $T$ is compared with all reference signatures $R \in \mathcal{R}$ and a signature verification score is calculated.
The signature $T$ is accepted as genuine if the signature verification score is below a certain threshold.
In the following subsections, the process of calculating the signature verification score is described in more detail.
This process is independent from the chosen dissimilarity score. 
In the following equations, $d(\cdot,\cdot)$ can stand for either $d_\text{GED}(\cdot,\cdot)$ or $d_\text{inkball}(\cdot,\cdot)$

\subsection{Reference-based Normalization}
It is expected that each user shows a different degree of variability in his/her signatures.
Based on the reference signatures of a given user, we want to predict how much variability we expect for this user and normalize the score accordingly.
This is done by applying a normalization with respect to the average dissimilarity among the reference signatures~\cite{Fischer2015RobustVerification}.
Specifically, each reference signature is compared against the other reference signatures and the average of the minima is calculated. Formally,
\begin{equation}
\delta(\mathcal{R}) = \avg_{R \in \mathcal{R}}\min_{S \in \mathcal{R} \setminus {R}}d(R, S)
\end{equation}
This score is used to normalize the dissimilarity scores of each user.
Formally, we define $\hat{d}(R, T, \mathcal{R})$ as the reference normalized score:
\begin{equation}
\hat{d}(R, T, \mathcal{R}) =  \frac{d(R, T)}{\delta(\mathcal{R})} \text{,}
\end{equation}
where $T$ is the questioned signature image, $R \in \mathcal{R}$ is a reference signature.

\subsection{Signature Verification Score}
The signature verification score $d(\mathcal{R}, T)$ is calculated given a set of reference signature images $\mathcal{R}$ and the questioned test signature image $T$.
Formally,
\begin{equation}
d(\mathcal{R},T) = \min_{R \in \mathcal{R}} \hat{d}(R,T,\mathcal{R}) = {\frac{\min_{R \in \mathcal{R}}{d(R, T)}}{\delta(\mathcal{R})}}
\end{equation}
The signature $T$ will be accepted if the minimum dissimilarity $d(\mathcal{R},T)$ between the references $\mathcal{R}$ and $T$ is below a certain threshold.

\subsection{Multiple Classifier System \label{sec:mcs}}
We define a linear combination of the two dissimilarity scores $d_\text{GED}$ and $d_\text{inkball}$ as a combined dissimilarity score $d_\text{MCS}$.
The idea is to leverage complementary aspects of the two dissimilarity measures to achieve better verification results using this multiple classifier system (MCS).
The dissimilarity scores are combined on an image pair basis, i.e.~the GED dissimilarity score between reference image $R$ and test image $T$ is combined with the inkball dissimilarity score between $R$ and $T$.
Both dissimilarity scores are z-score normalized based on all reference signatures before they are combined with each other. 
Formally,
\begin{equation}
\begin{split}
d_{\text{MCS},w}(\mathcal{R},T) = \min_{R \in \mathcal{R}} \Big( w \cdot \hat{d}^*_{\text{GED}}(R,T,\mathcal{R})
+ (1-w) \cdot \hat{d}^*_{\text{\old{inkball}}}(R,T,\mathcal{R}) \Big)
\end{split}
\end{equation}
where $w \in [0, 1]$ is a weighting factor, and
\begin{equation}
\hat{d}^*(R,T,\mathcal{R}) = \frac{\hat{d}(R,T,\mathcal{R}) - \mu_\mathcal{D}}{\sigma_\mathcal{D}} \text{,}
\end{equation}
where the mean $\mu_\mathcal{D}$ and the standard deviation $\sigma_\mathcal{D}$ are calculated over the set $\mathcal{D} = \{\mathcal{R}_1, \dots, \mathcal{R}_n\}$ of all references signature sets $\mathcal{R}_i$ of all $n$ users in the current dataset. 
Formally, the mean $\mu_\mathcal{D}$ is calculated as
\begin{equation}
\mu_\mathcal{D} = \avg_{\mathcal{R} \in \mathcal{D}} \Big( \avg_{R \in \mathcal{R}} \big( \min_{S \in \mathcal{R} \setminus {R}}\hat{d}(R,S,\mathcal{R}) \big) \Big)
\end{equation}
and the standard deviation $\sigma_\mathcal{D}$ is calculated as
\begin{equation}
\sigma_\mathcal{D} = \sqrt{\avg_{\mathcal{R} \in \mathcal{D}} \Big( \avg_{R \in \mathcal{R}} \big( \min_{S \in \mathcal{R} \setminus {R}}\hat{d}(R,S,\mathcal{R}) - \mu_\mathcal{D} \big)^2 \Big)}
\end{equation}

%

\section{Experimental Evaluation}\label{sec:exp}


\subsection{Datasets}

We use four publicly available datasets to evaluate the performance of our structural approaches. 
Table \ref{tab:datasets} gives an overview of the datasets, which are described in more detail in the following paragraphs.

\textbf{GPDSsynthetic-Offline} is a large dataset~\cite{Ferrer2015} of synthetic Western signatures. 
It is the replacement for the popular GPDS-960 dataset and its earlier variants, which are no longer available~\cite{GPDS2016website}. 
The new dataset consists of 4,000 synthetic users with 24~genuine signatures and 30~simulated forgeries for each user. 
All signatures have been generated with differently modeled pens at a simulated resolution of 600 dpi. 
We are using two subsets of this dataset:
\begin{itemize}
    \item \textbf{GPDS-last100:} Containing the last 100 users of the dataset (users 3901 to 4000). 
    \item \textbf{GPDS-75:} Containing the first 75 users of the dataset (users 1 to 75). 
\end{itemize}
The GPDS-last100 dataset is used as the training set for both structural methods. 
All parameters are tuned on this subset exclusively. 
While GPDS-75 is only used for testing and comparing against the state of the art.

\textbf{UTSig} is a rather new Persian signature dataset~\cite{Soleimani2016utsig}. 
It consists of 115~users with 27~genuine signatures, 3~opposite-hand signatures\footnote{The opposite-hand signatures are treated as forgeries as suggested by the authors of the dataset.}, and 42~forgeries for each user. 
The users have been instructed to sign within six differently sized bounding boxes to simulate different conditions. 
The resulting signatures have been scanned with 600 dpi. 
We use this dataset only for testing and comparing against the state of the art.

\textbf{MCYT-75} is a offline signature dataset within the MCYT baseline corpus~\cite{Ortega-Garcia2003,Fierrez-Aguilar2004}. 
It consists of 75 users with 15~genuine signatures and 15~forgeries for each user. 
The users signed in a $127\text{mm} \times 97\text{mm}$ box and each signature has been scanned at 600 dpi. 
We use this dataset only for testing and comparing against the state of the art.

The \textbf{CEDAR} dataset consists of 55 users~\cite{Kalera2004cedar} with 24~genuine signatures and 24~forgeries for each user. 
The users signed in a $50\text{mm} \times 50\text{mm}$ box and each signature has been scanned at 300 dpi. 
We use this dataset only for testing and comparing against the state of the art.

\begin{table}
\centering
\caption{Signature datasets \label{tab:datasets}}

\resizebox{0.9\linewidth}{!}{%
\begin{tabular}{@{}lcccccc@{}}
\toprule
\multirow{2}{*}{Name} & \multirow{2}{*}{Users} & \multirow{2}{*}{Genuine} & \multirow{2}{*}{Forgeries} & \multirow{2}{*}{dpi} & used for  & used for \\
     &  &  &  &  & tuning & testing  \\
 \midrule
GPDS-last100~\cite{Ferrer2015}          & 100   & 24 & 30 & 600 & x &   \\
GPDS-75~\cite{Ferrer2015}               & 75    & 24 & 30 & 600 &   & x \\
MCYT-75~\cite{Fierrez-Aguilar2004}      & 75    & 15 & 15 & 600 &   & x \\
UTSig~\cite{Soleimani2016utsig}         & 115   & 27 & 45 & 600 &   & x \\
CEDAR~\cite{Kalera2004cedar}            & 55    & 24 & 24 & 300 &   & x \\
\bottomrule
\end{tabular}
}
\end{table}

\subsection{Types of Forgeries}
We evaluate the performance of our classifiers to distinguish between genuine signatures and forgeries.
We are using two types of forgeries, which are common in the signature verification community:
\begin{itemize}
    \item \textbf{Skilled forgeries (SF):} 
    The target's genuine signature is known to the forger and usually, the forger has time to practice it.
    This often leads to forgeries that have high resemblance with their genuine counterpart.
    
    \item \textbf{Random forgeries (RF):} 
    Genuine signatures of other users are used in a brute force attack on the verification system. 
    Another reasoning is that the forgers use their own signatures since they have no knowledge about the target's signature.
    In our experiments, we are using one genuine signature from each other user as random forgeries.
\end{itemize}

\subsection{Number of References}
How many genuine signatures are used as references per users varies in literature even for the same dataset. 
We are labeling our results with $\text{R}x$ where $x$ is replaced with the number of references used, e.g. R10 means ten references are used for each user. 
We are always using the first $x$ genuine signature for each user. 
The remaining genuine signatures are used as positive samples for the evaluation.

\subsection{Evaluation Metrics}
\label{exp:metrics}

We evaluate the performance of our graph-based verification systems using the equal error rate (EER). 
The EER is the error rate when the false rejection rate (FRR) is equal to the false acceptance rate (FAR). 
The FRR refers to the percentage of genuine signatures that are rejected by the system and the FAR refers to the percentage of forgeries accepted by the system. 
In order to determine FRR and FAR directly, we have to decide on a decision threshold (see Section \ref{exp:threshold}). 
We distinguish EER and FAR based on the type of forgeries. 
For skilled and random forgeries, we call them $\text{EER}_\text{SF}$/$\text{FAR}_\text{SF}$ and $\text{EER}_\text{RF}$/$\text{FAR}_\text{RF}$, respectively. 
We calculate an average error rate (AER) using the following equation.
\begin{equation}
    \text{AER}_\text{SF} = \frac{\text{FRR} + \text{FAR}_\text{SF}}{2}
\end{equation}
Furthermore, we calculate two types of EER, i.e. the global EER and the user-specific EER.
The global EER ($\text{EER}^\text{global}_\text{RF}$ and $\text{EER}^\text{global}_\text{SF}$) is calculated by using the same (global) threshold for all users.
The user-specific EER $\text{EER}^\text{user}_\text{RF}$ and $\text{EER}^\text{user}_\text{SF}$) is the average of all individual EERs calculated per user.


\subsection{Parameter Tuning}
We tune the parameters of our graph-based methods on our tuning set GPDS-last100. 
We are using ten references per user (R10) and optimize the parameters with respect to $\text{EER}_\text{SF}$. 
We also measure the runtime per comparison since we expect a significant increase of the computation time with growing graph sizes.

\subsubsection{Graph Edit Distance}
In a first step, we create several sets of graphs using different values for $D_\text{GED}$. 
Table \ref{tab:keypoint-stats} shows the minimum, median, average, and maximum number of nodes in the graphs for a given $D_\text{GED}$. 
The number of nodes is increasing as expected when lowering $D_\text{GED}$.

\begin{table}
\centering
\caption{Nodes in keypoint graph based on $D_\text{GED}$ for first 10 genuine per user of GPDS-last100 dataset. \label{tab:keypoint-stats}}

\resizebox{0.6\linewidth}{!}{%
\begin{tabular}{c|cccc}
\toprule
$D_\text{GED}$ & minimum & median & average & maximum\\
 \midrule
 100 &  9 &  42 &  45 & 120  \\
  50 & 15 &  70 &  73 & 194  \\
  45 & 16 &  76 &  79 & 214  \\
  40 & 17 &  84 &  87 & 236  \\
  35 & 18 &  93 &  98 & 265  \\
  30 & 19 & 106 & 111 & 307  \\
  25 & 23 & 125 & 130 & 355  \\
  20 & 25 & 152 & 159 & 439  \\
  15 & 31 & 198 & 206 & 568  \\
  10 & 42 & 286 & 299 & 835  \\
\bottomrule
\end{tabular}
}

\end{table}


The first question we want to answer is whether BP or HED performs better for the task of signature verification. 
For this experiment, we use the keypoint graphs for $D_\text{GED} \in \{25, 50, 100\}$. 
We optimize the cost function, i.e. $c_\text{node}$ and $c_\text{edge}$, individually using a grid search for each $D_\text{GED}$ and the approximations $\bp$ and $\hed$. 
In Table \ref{tab:BPvsHED}, \af{we report} the results and runtime\footnote{Runtime is with respect to a Java implementation and AMD Opteron 2354 nodes with 2.2 GHz CPU.} for GPDS-last100 R10. 
The EER results of $\hed$ are very similar to $\bp$, while being much faster. 
This speed-up is particularly significant when using graph representations with more nodes. 
Due to the speed-up and the similar performance in EER, we continue the following experiments with the $\hed$ approximation. 

We extend the grid search to smaller $D_\text{GED}$ using the $\hed$ approximation. 
The best $\text{EER}_\text{SF}$ results on GPDS-last100 R10 is achieved by $D_\text{GED}=25$, $c_\text{node}=12.5$, and $c_\text{edge}=200$.
We use this setup as our proposed GED approach.

\begin{table}
\centering
\caption{Comparison of GED approximations on GPDS-last100 R10 using $\text{EER}_\text{SF}$ and average runtime per comparison. 
The overall best result is shown in bold font.\label{tab:BPvsHED}}

\resizebox{0.6\linewidth}{!}{%
\begin{tabular}{cc|cc|cc}
\toprule
$D_\text{GED}$ & Approx. & $\text{EER}_\text{SF}$ & runtime & $c_\text{node}$ & $c_\text{edge}$ \\
 \midrule
\multirow{2}{*}{25}  & $\bp$  &  7.87 & 1029 ms & 25   &  25  \\
                     & $\hed$ &  \textbf{7.47} &  113 ms & 12.5 & 200  \\
\midrule
\multirow{2}{*}{50}  & $\bp$  &  7.77 &  175 ms & 12.5 &  50  \\
                     & $\hed$ &  8.27 &   38 ms & 20   & 150  \\
\midrule
\multirow{2}{*}{100} & $\bp$  & 10.53 &   51 ms & 50   &  50  \\
                     & $\hed$ & 11.03 &   17 ms & 50   & 100  \\
\bottomrule
\end{tabular}
}



\end{table}

\subsubsection{Inkball}
For the inkball approach, we again started with creating models using different values for $D_\text{inkball}$. 
Table \ref{tab:inkball-stats} shows the minimum, median, average, and maximum number of nodes in the inkball models for a given $D_\text{inkball}$. 

\begin{table}
\centering
\caption{Nodes in inkball model based on $D_\text{inkball}$ for the first 10 genuine per user of the GPDS-last100 dataset. \label{tab:inkball-stats}}

\resizebox{0.6\linewidth}{!}{%
\begin{tabular}{c|cccc}
\toprule
$D_\text{inkball}$ & minimum & median & average & maximum\\
 \midrule
 100 &   8 &   33 &   35 &   82  \\
  50 &  17 &   56 &   59 &  133  \\
  40 &  20 &   69 &   72 &  166  \\
  32 &  26 &   86 &   89 &  202  \\
  25 &  35 &  110 &  114 &  261  \\
  16 &  52 &  170 &  177 &  398  \\
  10 &  76 &  270 &  279 &  626  \\
   8 &  90 &  334 &  347 &  784  \\
   6 & 119 &  438 &  454 & 1034  \\
   4 & 174 &  650 &  677 & 1544  \\
\bottomrule
\end{tabular}
}
\end{table}

To compare the novel augmented inkball models with the normal inkball models, we test both approaches with different values for $D_\text{inkball} \in \{4,6,8,16,32\}$. 
The results and runtime\footnote{Runtime is with respect to a C implementation and AMD Opteron 2354 nodes with 2.2 GHz CPU.} for GPDS-last100 R10 are shown in Table \ref{tab:PSMvsAugmented}. 
The augmented inkball matching shows consistently better results than the standard inkball matching while being about 50\% slower considering the same $D_\text{inkball}$, i.e. the same number of inkballs. 
We can also see that the inkball matching is significantly slower than the HED approximation approach. 
This is in part due to the smaller values for $D_\text{inkball}$ and therefore the much larger models.
Additionally, the models are matched against a skeleton image, which contains more matching points than a keypoint graph with $D_\text{GED}>1$.
But it is probably also due to the overall higher computational complexity of the inkball matching. 

\new{Note that instead of increasing $D_\text{inkball}$, the resolution of the signature images can also be lowered to speed up computation.
This reduces the runtime significantly, however, it may also lead to a decrease in verification performance due to the loss of details in the signature image.}

As shown in Table \ref{tab:PSMvsAugmented}, the best results on our tuning set are achieved using the augmented inkball matching using $D_\text{inkball} = 6$, $\text{mdsq}=64$, and $\text{angwgt}=64$. 
We use this configuration as our proposed inkball approach.

\begin{table}
\centering
\caption{Augmented inkball compared to normal inkball models on GPDS-last100 R10 using $\text{EER}_\text{SF}$ and average runtime per comparison.
The overall best result is shown in bold font.\label{tab:PSMvsAugmented}}

\resizebox{0.7\linewidth}{!}{%
\begin{tabular}{cc|cc|cc}
\toprule
$D_\text{inkball}$ & Approach & $\text{EER}_\text{SF}$ & runtime & mdsq & angwgt \\
 \midrule
\multirow{2}{*}{32}  & normal    & 12.33 &  5.65 s  &  96 & - \\
                     & augmented & 11.20 &  8.58 s  & 128 & 64 \\
\midrule
\multirow{2}{*}{16}  & normal    &  9.97 & 11.19 s  &  96 & - \\
                     & augmented &  9.27 & 16.86 s  &  96 & 64 \\
\midrule
\multirow{2}{*}{8}   & normal    &  9.83 & 22.53 s  &  64 & - \\
                     & augmented &  8.63 & 33.53 s  & 128 & 64 \\
\midrule
\multirow{2}{*}{6}   & normal    &  9.63 & 28.80 s  &  32 & - \\
                     & augmented &  \textbf{8.27} & 44.37 s  &  64 & 64 \\
\midrule
\multirow{2}{*}{4}   & normal    &  9.53 & 42.83 s  &  16 & - \\
                     & augmented &  8.33 & 66.07 s  &  32 & 64 \\
\bottomrule
\end{tabular}
}
\end{table}

\subsubsection{Multi-Classifier System}
We combine the GED approach with the inkball approach using the MCS approach described in Section \ref{sec:mcs}.
To find the optimal weight for this combination, we evaluate different weights $w \in \{0.0, 0.1, \dots, 0.9, 1.0\}$ on the GPDS-last100 R10 with respect to $\text{EER}_\text{SF}$.
The results are shown in Table \ref{tab:MCSweight}.
The best result is achieved when applying equal weights to both methods, i.e. $w=0.5$.

\begin{table}
\centering
\caption{$\text{EER}^\text{global}_\text{SF}$ results for a MCS with different weights $w$ on GPDS-last100 R10.
The overall best result is shown in bold font. \label{tab:MCSweight}}

\resizebox{0.82\linewidth}{!}{%
\begin{tabular}{c|ccccccccccc}
\toprule
$w$                    & 0.0  & 0.1  & 0.2  & 0.3  & 0.4  & \textbf{0.5}  & 0.6  & 0.7  & 0.8  & 0.9  & 1.0  \\
 \midrule
$\text{EER}^\text{global}_\text{SF}$ & 8.3 & 7.2 & 6.8 & 6.3 & 6.0 & \textbf{5.5} & 5.7 & 6.0 & 6.8 & 7.2 & 7.5 \\
\bottomrule
\end{tabular}
}
\end{table}


The results of our proposed MCS system clearly outperforms our proposed individual methods as shown in Table~\ref{tab:MCSvsSingle}.

\begin{table*}
\centering
\caption{Comparing proposed MCS system with proposed GED and proposed inkball on the development set (GPDS-last100 R10). 
The best results are highlighted in bold font. \label{tab:MCSvsSingle}}

\begin{adjustbox}{width=0.45\textwidth, keepaspectratio}
\begin{tabular}{l|cc}
\toprule
System & $\text{EER}^\text{user}_\text{SF}$  & $\text{EER}^\text{global}_\text{SF}$ \\

\midrule
\textbf{Proposed} GED       & 6.03 & 7.47 \\ 
\textbf{Proposed} Inkball   & 5.33 & 8.27 \\ 
\textbf{Proposed} MCS       & \textbf{3.30} & \textbf{5.50} \\ 

\bottomrule
\end{tabular}
\end{adjustbox}
\end{table*}

\subsection{Threshold Selection}
\label{exp:threshold}
A decision threshold is required to calculate the $\text{FRR}$, $\text{FAR}_\text{SF}$, $\text{FAR}_\text{RF}$, and $\text{AER}$.
The threshold that leads to the $\text{EER}^\text{global}_\text{SF}$ on the GPDS-last100 dataset is used as the decision threshold for each proposed system.
Since we are dealing with three proposed systems, namely \emph{proposed GED}, \emph{proposed Inkball}, \emph{proposed MCS}, we end up with three decision thresholds. 
The decision threshold for a specific proposed system is employed for all users, test sets, and experiments (skilled and random forgeries).
Choosing the decision threshold this way is a simple approach.
However, it relies on a good tuning dataset to find a good decision threshold.
The decision threshold is only applied for FRR, FAR, and AER. 
The EER calculation is not affected by this decision threshold.

\subsection{Test Results}
\begin{figure}[t!]
\centering
\subfloat[GPDS-75\label{fig:det-curves-gpds}]{
    \includegraphics[width=.45\textwidth,trim={0mm 0mm 0mm 0mm},clip]{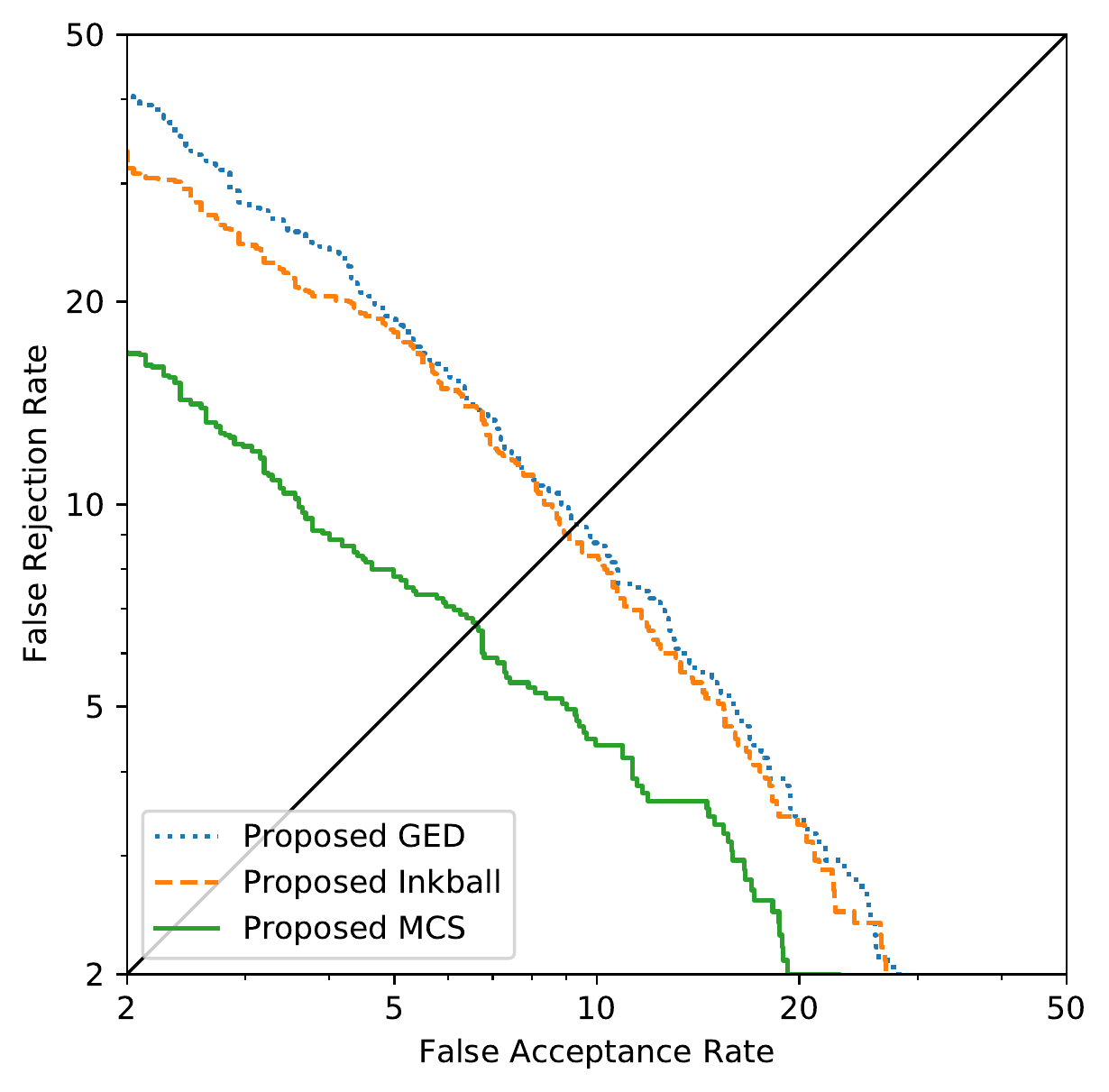}
}
\subfloat[UTSig\label{fig:det-curves-utsig}]{
    \includegraphics[width=.45\textwidth,trim={0mm 0mm 0mm 0mm},clip]{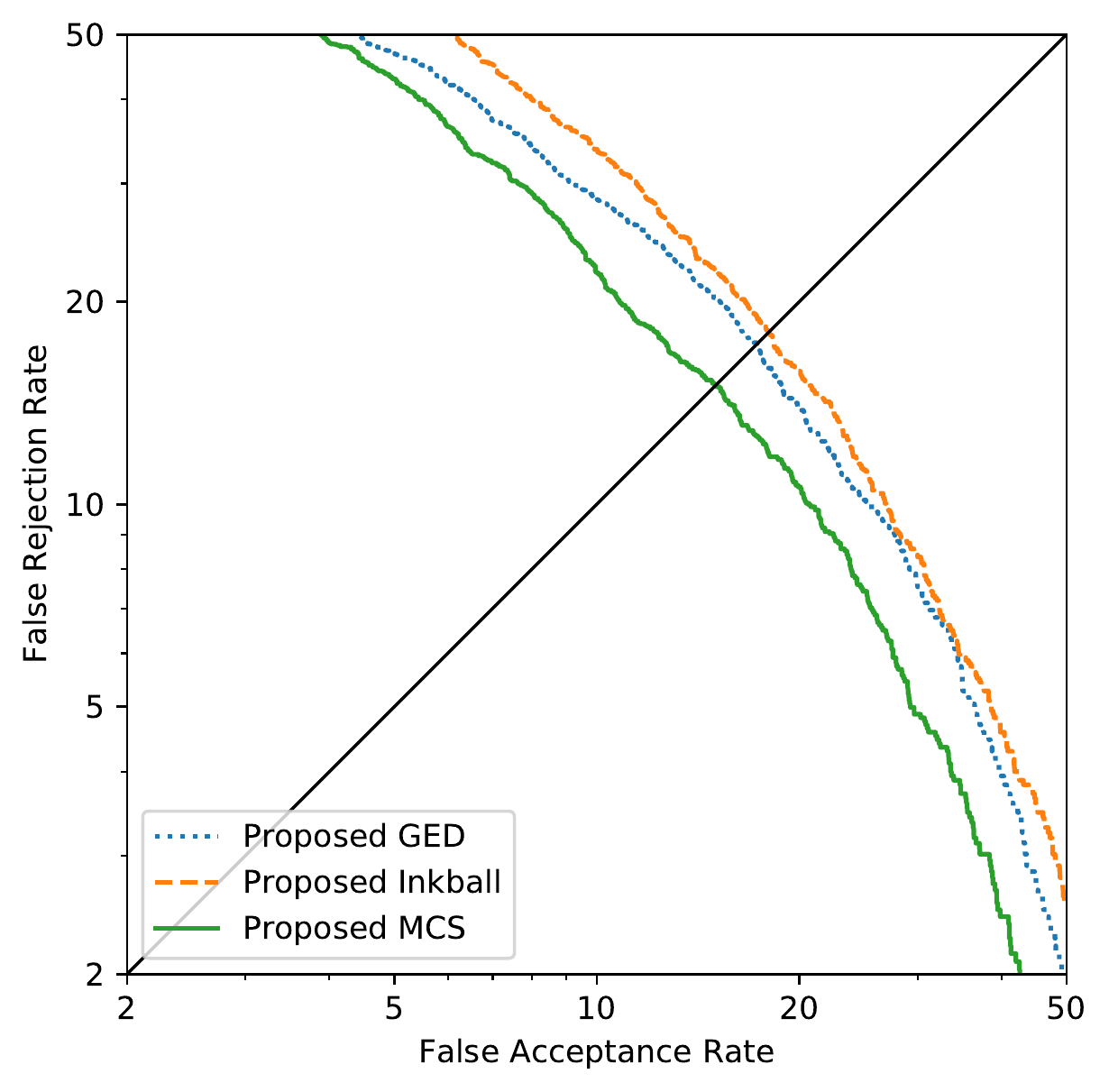}
} \quad
\subfloat[MCYT-75\label{fig:det-curves-mcyt}]{
    \includegraphics[width=.45\textwidth,trim={0mm 0mm 0mm 0mm},clip]{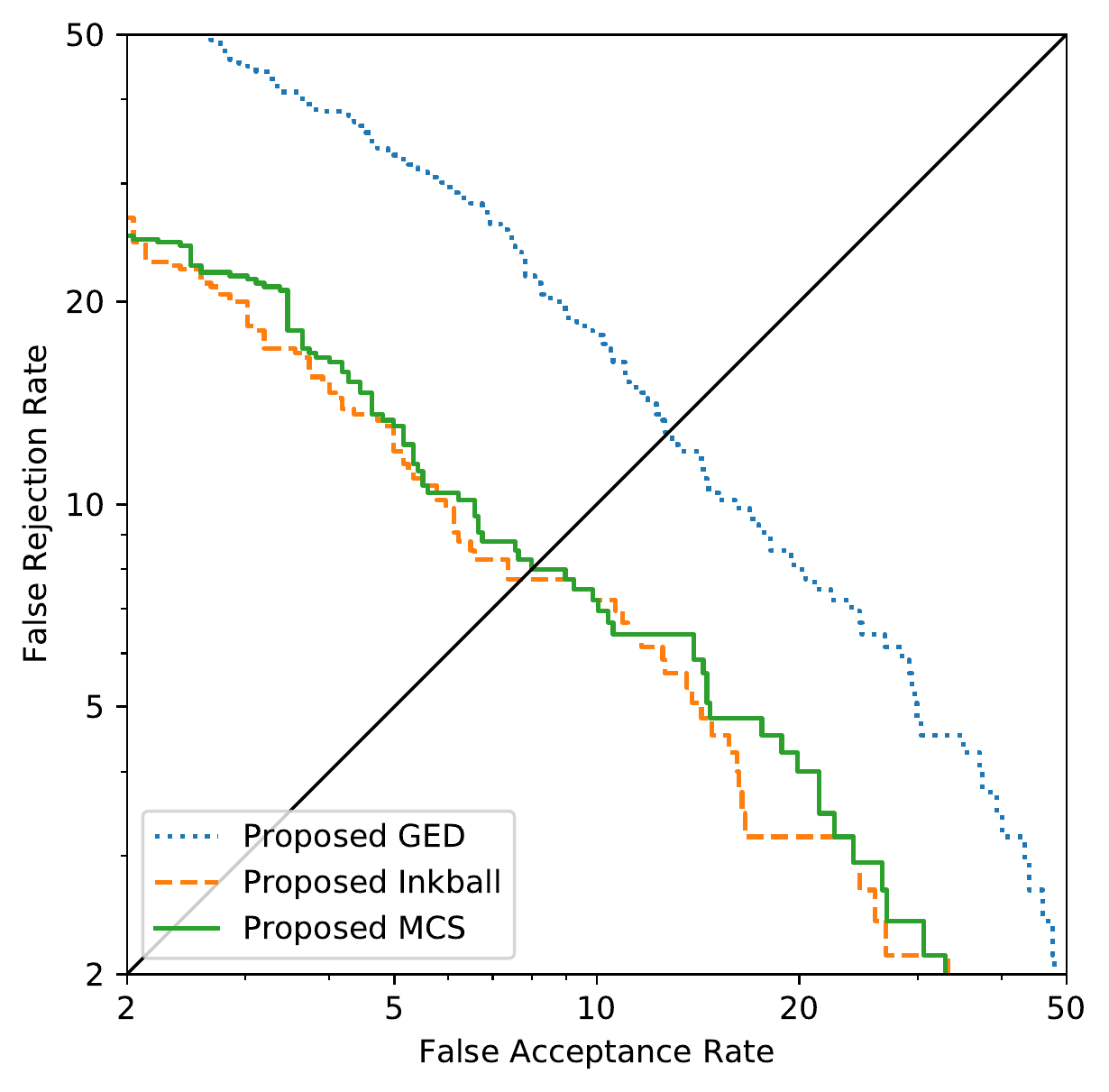}
}
\subfloat[CEDAR\label{fig:det-curves-cedar}]{
    \includegraphics[width=.45\textwidth,trim={0mm 0mm 0mm 0mm},clip]{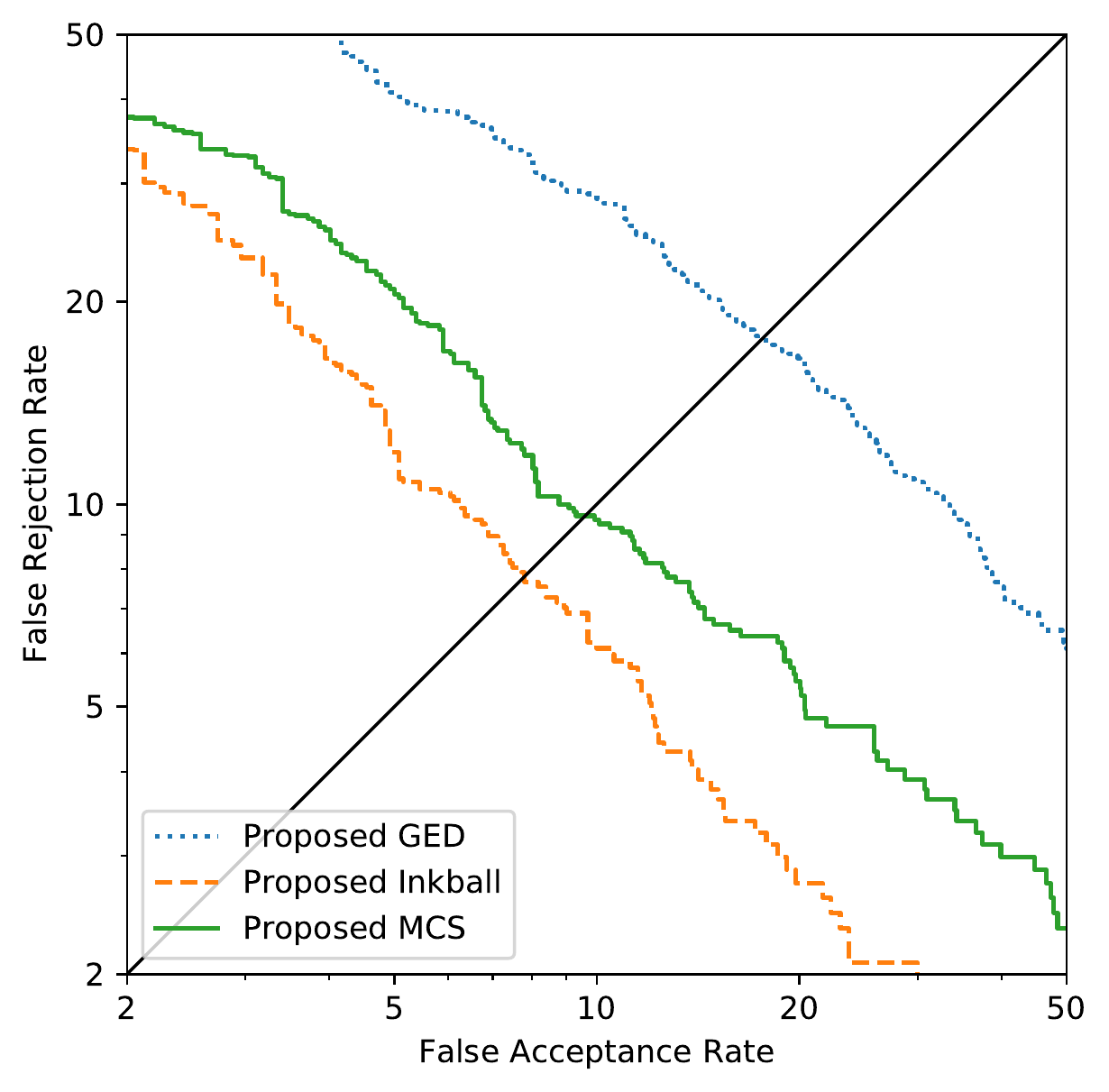}
}
\caption{DET curves. \label{fig:det-curves}}
\end{figure}

The three proposed systems, which have been tuned on the GPDS-last100 dataset, are applied without further adaptation on the four test sets (GPDS-75, MCYT-75, UTSig, and CEDAR).
Several different evaluation protocols have been used in past publications making meaningful comparison often not trivial.
We adopt an evaluation protocol that has been utilized in recent state-of-the-art publications and measure the performance with regard to eight evaluation metrics (see Section \ref{exp:metrics}).
The results are compared against several published state-of-the-art results that have been obtained using the same (or almost identical) evaluation protocol.
Tables \ref{tab:comparison-gpds} to \ref{tab:comparison-cedar} present the results and Fig.~\ref{fig:det-curves} shows the DET curves for the results of the proposed systems.

First, we compare the three proposed systems with each other.
The proposed MCS achieves better results than the individual proposed systems on two of the four test sets (GPDS-75 and UTSig) and better results for random forgeries on the MCYT-75.
The proposed inkball system obtains better results than proposed GED and MCS on CEDAR and on MCYT-75 for skilled forgeries.
Overall, the proposed MCS achieves the best results on the four datasets.

On GPDS-75 (Table \ref{tab:comparison-gpds}), the proposed MCS produces excellent results with achieving the best result with regards to five of the eight evaluation metrics.
Only, the results on random forgeries are a little weaker.
The used training data is similar to this test set and thus, particularly useful.
Overall, these results demonstrate the excellent performance that can be achieved when using specific training data.

On UTSig (Table \ref{tab:comparison-utsig}), the proposed MCS obtains the best results with regard to five of the eight evaluation metrics.
This is particularly impressive since this test set consists of Persian signatures, while the proposed systems have been trained on synthetic Western signatures.
These results highlight that the proposed system works even on unseen scripts.

On MCYT-75 (Table \ref{tab:comparison-mcyt}), the proposed inkball system obtains top-3 results with regard to seven of the eight evaluation metrics (it achieves the best result according to three metrics).
As mentioned before, the proposed MCS achieves slightly worse results than the proposed inkball system.
Even though our results are very good, the best-reported results are significantly better in some of the metrics.
This might be caused by the absence of dataset-specific training.
However, overall the proposed method performs well on this specific signature dataset, especially considering that our method is trained on synthetic data only.

On CEDAR (Table \ref{tab:comparison-cedar}), the proposed inkball system achieves top-2 results with regard to six of the eight evaluation metrics.
However, the number of publications that follow the same evaluation protocol on this test set is limited, which is reducing the number of published results we can use for comparison.
Overall, the state of the art achieves better results than our proposed systems.
Especially, the proposed GED system obtains poor results affecting also the performance of our proposed MCS.
It is possible that this is due to the lower resolution of the signature images in this \af{test set~(see Table \ref{tab:datasets})}.
Nonetheless, the performance of the proposed inkball system in the user-specific metrics ($\text{EER}^\text{user}_\text{RF}$ and $\text{EER}^\text{user}_\text{SF}$) indicates that our structural approach has potential on this test set.

To sum up, the results show that structural approaches have a high potential for signature verification, especially when dealing with skilled forgeries.
The proposed inkball system shows a more consistent performance across all four datasets when compared to the proposed GED system. 
However, the long runtime of the inkball approach might be an issue for a real-world application.
The performance of the proposed systems when encountering random forgeries is not as strong as the state of the art.
As proposed by~\cite{maergner2018ssspr}, a combination with a neural network should \af{further improve} the performance of the structural approach on random forgeries. 
The results also show that the difference between the user-specific and the global EER is quite large.
This indicates that the proposed approaches could benefit from improved user adaptation in the future.
Overall, the results are quite remarkable, especially when considering that the proposed systems are applied on four different test sets without any further adaptation while being trained on synthetic signatures only.

\begin{table*}
\centering
\caption{GPDS-75 dataset: Comparison with other published methods. 
The best result is highlighted in bold font and the top three results are marked with numbers.\label{tab:comparison-gpds}}

\begin{adjustbox}{width=\textwidth, totalheight=\textheight-2\baselineskip, keepaspectratio}
\begin{tabular}{lcllllllll}
\toprule
System & \#Refs & FRR & $\text{FAR}_\text{RF}$ & $\text{EER}^\text{user}_\text{RF}$ & $\text{EER}^\text{global}_\text{RF}$ & $\text{FAR}_\text{SF}$ & $\text{AER}_\text{SF}$  & $\text{EER}^\text{user}_\text{SF}$  & $\text{EER}^\text{global}_\text{SF}$\\
\midrule
GPDS website~, 2016~\cite{GPDS2016website}          & 10 & -       & -       & -       & 0.76 (2)&\x-       &  -       & -       & 16.01    \\ 
Soleimani et al., 2016~\cite{Soleimani2016patreclet}& 10 & 6.51*(3)& \textbf{0.11}*(1)& -       & 1.08 (3)& 18.23*   & 12.37*   & -       & 12.83    \\ 
Maergner et al., 2018~\cite{maergner2018ssspr}      & 10 & -       & -       & -       & \textbf{0.56} (1)&\x-       &  -       & -       &\x7.24 (3)\\ 
Maergner et al., 2018~\cite{maergner2018icfhr}      & 10 & -       & -       & -       & 2.05    &\x-       &  -       & -       &\x6.84 (2)\\ 
Narwade et al., 2018~\cite{Narwade2018}             & 12 & \textbf{3.51}*(1)& -       & -       & -       & 13.91*   &\x8.71*(2)& -       &\x-       \\ 
&&&&&&&&&\\
\textbf{Proposed GED}                               & 10 & 8.00    & 1.21 (3)& 1.53 (3)& 3.89    & 10.76 (3)&\x9.38    & 6.67 (2)&\x9.33    \\
\textbf{Proposed Inkball}                           & 10 & 8.76    & 1.42    & 1.42 (2)& 3.42    &\x9.11 (2)&\x8.94 (3)& 6.71 (3)&\x9.02    \\
\textbf{Proposed MCS}                               & 10 & 6.38 (2)& 0.25 (2)& \textbf{0.59} (1)& 2.27    &\x\textbf{6.76} (1)&\x\textbf{6.57} (1)& \textbf{4.67} (1)&\x\textbf{6.62} (1)\\
\bottomrule
\multicolumn{10}{l}{\footnotesize{*: The starred numbers have been calculated for 2500 users (Soleimani et al. \citep{Soleimani2016patreclet}) and 90 users (Narwade et al. \citep{Narwade2018}).}} \\
\multicolumn{10}{l}{\footnotesize{\phantom{*:} However, results for 75 users should be similar since this dataset is quite stable for different user counts (see results on GPDS website \citep{GPDS2016website}).}} \\
\end{tabular}
\end{adjustbox}
\end{table*}

\begin{table*}
\centering
\caption{UTSig dataset: Comparison with other published methods. 
The best result is highlighted in bold font and the top three results are marked with numbers.\label{tab:comparison-utsig}}

\begin{adjustbox}{width=\textwidth, totalheight=\textheight-2\baselineskip, keepaspectratio}
\begin{tabular}{lcllllllll}
\toprule
System & \#Refs & FRR & $\text{FAR}_\text{RF}$ & $\text{EER}^\text{user}_\text{RF}$ & $\text{EER}^\text{global}_\text{RF}$ & $\text{FAR}_\text{SF}$ & $\text{AER}_\text{SF}$  & $\text{EER}^\text{user}_\text{SF}$  & $\text{EER}^\text{global}_\text{SF}$\\
\midrule
Soleimani et al., 2016~\cite{Soleimani2016utsig}     & 12 & 39.27    & 0.08 (3)& -       & -       & 21.29    & 30.28*   &\x-       & 29.71    \\ 
Soleimani et al., 2016~\cite{Soleimani2016patreclet} & 12 & 18.96    & \textbf{0.00} (1)& -       & -       & 16.15 (2)& 17.56*   &\x-       & 17.45    \\ 
Soleimani et al., 2016~\cite{Soleimani2016ICCKE}     & 12 & 16.34    & 0.01 (2)& -       & -       & \textbf{15.69} (1)& 16.02*(2)&\x-       & 16.00 (2)\\ 
Narwade et al., 2018~\cite{Narwade2018}             &  9 &\x\textbf{7.41} (1)& -       & -       & -       & 24.95    & 16.18 (3)&\x-       &\x-       \\ 
&&&&&&&&&\\
\textbf{Proposed GED}                               & 12 & 17.80    & 0.85    & 2.14 (2)& 4.00 (2)& 16.75 (3)& 17.28    & 14.11 (2)& 17.33 (3)\\
\textbf{Proposed Inkball}                           & 12 & 14.14 (3)& 2.11    & 2.38 (3)& 5.02 (3)& 22.26    & 18.20    & 14.61 (3)& 18.03    \\
\textbf{Proposed MCS}                               & 12 &\x8.75 (2)& 0.98    & \textbf{1.11} (1)& \textbf{2.97} (1)& 22.98    & \textbf{15.86} (1)& \textbf{11.75} (1)& \textbf{15.01} (1)\\
\bottomrule
\end{tabular}
\end{adjustbox}
\end{table*}

\begin{table*}
\centering
\caption{MCYT-75 dataset: Comparison with other published methods. 
The best result is highlighted in bold font and the top three results are marked with numbers.\label{tab:comparison-mcyt}}

\begin{adjustbox}{width=\textwidth, totalheight=\textheight-2\baselineskip, keepaspectratio}
\begin{tabular}{lcllllllll}
\toprule
System & \#Refs & FRR & $\text{FAR}_\text{RF}$ & $\text{EER}^\text{user}_\text{RF}$ & $\text{EER}^\text{global}_\text{RF}$ & $\text{FAR}_\text{SF}$ & $\text{AER}_\text{SF}$  & $\text{EER}^\text{user}_\text{SF}$  & $\text{EER}^\text{global}_\text{SF}$\\
\midrule
Fierrez-Aguilar et al., 2004~\cite{Fierrez-Aguilar2004}             & 10 &\x-       & -       & 1.14*   & -       &\x-       &\x-       &\x9.28*   &\x-       \\ 
Alonso-Fernandez et al., 2007~\cite{Alonso-Fernandez2007Automatic}  & 10 &\x-       & -       & 7.26    & -       &\x-       &\x-       & 22.13    &\x-       \\ 
Gilperez et al., 2008~\cite{Gilperez2008Off-lineFeatures}           & 10 &\x-       & -       & 1.18    & -       &\x-       &\x-       &\x6.44    &\x-       \\ 
Vargas et al., 2011~\cite{Vargas2011}                               & 10 & 12.61    & 1.53    & -       & 2.20    &\x\textbf{7.53} (1)& 10.07*   &\x-       &\x8.80    \\ 
Ooi et al., 2016~\cite{Ooi2016}                                     & 10 &\x-       & -       & -       & -       &\x-       &\x-       &\x-       &\x9.87    \\ 
Soleimani et al., 2016~\cite{Soleimani2016patreclet}                & 10 &\x6.13 (2)& \textbf{0.00} (1)& -       & 0.37 (2)& 12.71 (2)&\x9.42*(2)&\x-       &\x9.86    \\ 
Hafemann et al., 2018~\cite{Hafemann2018}                           & 10 &\x-       & -       & \textbf{0.03} (1)& \textbf{0.19} (1)&\x-       &\x-       &\x-       &\x\textbf{3.64} (1)\\ 
Maergner et al., 2018~\cite{maergner2018ssspr}                      & 10 &\x-       & -       & 0.25 (2)& 0.79 (3)&\x-       &\x-       & 10.13    & 11.11    \\ 
Maergner et al., 2018~\cite{maergner2018icfhr}                      & 10 &\x-       & -       & 0.52    & 1.24    &\x-       &\x-       &\x5.78 (3)&\x8.71    \\ 
Narwade et al., 2018~\cite{Narwade2018}                             & 10 &\x-       & -       & -       & -       &\x-       &\x-       &\x-       &\x9.26    \\ 
&&&&&&&&&\\
\textbf{Proposed GED}                                               & 10 &\x7.20    & 1.78    & 1.39    & 3.87    & 22.84    & 15.02    &\x8.36    & 12.71    \\
\textbf{Proposed Inkball}                                           & 10 &\x\textbf{5.60} (1)& 0.92 (3)& 0.29 (3)& 2.70    & 12.89 (3)&\x\textbf{9.24} (1)&\x\textbf{3.02} (1)&\x7.73 (2)\\ 
\textbf{Proposed MCS}                                               & 10 &\x6.40 (3)& 0.43 (2)& 0.29 (3)& 1.91    & 13.69    & 10.04 (3)&\x3.47 (2)&\x8.00 (3)\\ 
\bottomrule
\end{tabular}
\end{adjustbox}
\end{table*}

\begin{table*}
\centering
\caption{CEDAR dataset: Comparison with other published methods. 
The best result is highlighted in bold font and the top three results are marked with numbers.\label{tab:comparison-cedar}}

\begin{adjustbox}{width=\textwidth, totalheight=\textheight-2\baselineskip, keepaspectratio}
\begin{tabular}{lcllllllll}
\toprule
System & \#Refs & FRR & $\text{FAR}_\text{RF}$ & $\text{EER}^\text{user}_\text{RF}$ & $\text{EER}^\text{global}_\text{RF}$ & $\text{FAR}_\text{SF}$ & $\text{AER}_\text{SF}$  & $\text{EER}^\text{user}_\text{SF}$  & $\text{EER}^\text{global}_\text{SF}$\\
\midrule
Chen et al., 2006~\cite{Chen2006}               & 16 &\x\textbf{7.70} (1)& -       & -       & -       &\x8.20    &\x\textbf{7.95}*(1)&\x-       &\x-         \\ 
Bharathi and Shekar, 2013~\cite{Bharathi2013}   & 12 &\x9.36 (2)& -       & -       & -       &\x7.84 (2)&\x8.60*(2)&\x-       &\x-         \\ 
Hafemann et al., 2018~\cite{Hafemann2018}       & 10 &\x-       & -       & 0.37 (2)& \textbf{1.14} (1)&\x-       &  -       &\x-       &\x\textbf{3.60} (1)  \\ 
&&&&&&&&&\\

\textbf{Proposed GED}                           & 10 & 18.70    & 0.51 (3)& 2.09    & 5.93    & 16.36    & 17.53    & 12.20 (3)& 17.50      \\
\textbf{Proposed Inkball}                       & 10 & 16.88    & \textbf{0.00} (1)& \textbf{0.24} (1)& 2.05 (2)&\x\textbf{3.94} (1)& 10.41    &\x\textbf{4.17} (1)&\x7.80 (2)  \\ 
\textbf{Proposed MCS}                           & 10 & 10.39 (3)& 0.03 (2)& 0.61 (3)& 2.96 (3)&\x8.18 (3)&\x9.29 (3)&\x5.76 (2)&\x9.55 (3)  \\ 
\bottomrule
\end{tabular}
\end{adjustbox}
\end{table*}

%

\section{Conclusions and Outlook}\label{sec:concl}
In this paper, two structural methods for signature verification are investigated.
It is shown that the graph edit distance based approach can be speeded up significantly while maintaining verification accuracy by using the Hausdorff edit distance as the approximation of the graph edit distance.
A novel augmented inkball matching that considers angular information is introduced, which leads to significant improvements in verification accuracy.
\af{While} the runtime of the inkball matching is certainly an issue that needs to be addressed for a real-world application, the verification results are impressive.
Additionally, our experiments show that the verification accuracy can be improved even further by combining the two structural methods.
Overall, the proposed structural approach achieves excellent performance on four publicly available signature verification datasets.
On two datasets, the proposed combined approach achieves the lowest EER on skilled forgeries compared to previously published results that have applied the same evaluation protocol.

Several future lines of research can be pursued to improve the proposed signature verification system.
First, the user-adaptation might be improved by modeling the signature stability of each user more closely using our structural models based on the reference signatures.
It is challenging to model the signature stability based on only a small number of reference signatures, but we believe that powerful structural representations offer a promising way to achieve this goal.
Finally, the proposed structural approaches would probably benefit from the complementary perspective of a statistical approach like convolutional neural networks, making the biometric authentication more robust.

\section*{\nrh{Acknowledgement}}
This work has been supported by the Swiss National Science Foundation project 200021\_162852.


\section*{\refname}
\begingroup
\renewcommand{\section}[2]{}%
\bibliographystyle{elsarticle-num}
\bibliography{biblio}
\endgroup

\end{document}